\documentclass[conference]{IEEEtran}
\IEEEoverridecommandlockouts
\usepackage{cite}
\usepackage{amsmath,amssymb,amsfonts}
\usepackage{algorithmic}
\usepackage{graphicx}
\if CLASSOPTIONcompsoc
    \usepackage[caption=false, font=normalsize, labelfont=sf, textfont=sf]{subfig}
\else
\usepackage[caption=false, font=footnotesize]{subfig}
\fi
\usepackage{textcomp}
\usepackage{xcolor}
\usepackage{lipsum}
\usepackage{makecell}
\def\BibTeX{{\rm B\kern-.05em{\sc i\kern-.025em b}\kern-.08em
    T\kern-.1667em\lower.7ex\hbox{E}\kern-.125emX}}

\usepackage[group-separator={,}]{siunitx}

\allowdisplaybreaks

\begin{document}

\title{Robust Constrained Multi-objective Evolutionary Algorithm based on Polynomial Chaos Expansion for Trajectory Optimization
}

\author{\IEEEauthorblockN{Yuji Takubo}
\IEEEauthorblockA{\textit{School of Aerospace Engineering} \\
\textit{Georgia Institute of Technology}\\
Atlanta, GA, USA \\
yujitakubo@gatech.edu}
\and
\IEEEauthorblockN{Masahiro Kanazaki}
\IEEEauthorblockA{\textit{Graduate School of Systems Design} \\
\textit{Department of Aerospace Engineering}\\
\textit{Tokyo Metropolitan University},
Tokyo, Japan \\
kana@tmu.ac.jp}
}

\maketitle

\begin{abstract}
An integrated optimization method based on the constrained multi-objective evolutionary algorithm (MOEA) and non-intrusive polynomial chaos expansion (PCE) is proposed, which solves robust multi-objective optimization problems under time-series dynamics.
The constraints in such problems are difficult to handle, not only because the number of the dynamic constraints is multiplied by the discretized time steps but also because each of them is probabilistic.
The proposed method rewrites a robust formulation into a deterministic problem via the PCE, and then sequentially processes the generated constraints in population generation, trajectory generation, and evaluation by the MOEA. As a case study, the landing trajectory design of supersonic transport (SST) with wind uncertainty is optimized. Results demonstrate the quantitative influence of the constraint values over the optimized solution sets and corresponding trajectories, proposing robust flight controls. 
\end{abstract}

\begin{IEEEkeywords}
Robust Optimization, Multi-objective Optimization, Evolutionary Algorithm, Polynomial Chaos Expansion, Trajectory Optimization
\end{IEEEkeywords}

\section{Introduction}
Uncertainty in optimization is critical for solving real-world problems, however the integration of stochastic aspects makes the problems drastically challenging compared to nominal ones. In spite of extensive efforts to treat probabilistic expressions deterministically, much less research on multi-objective problems can be found compared to single-objective cases. 

Furthermore, the introduction of a dynamical system exacerbates the complexity, as the number of the constraints that have to be addressed, which includes not only the dynamics itself but also the state or control constraints over the time horizon, is multiplied by the discretized time steps. State-of-the-art constraint-handling techniques \cite{coello2002theoretical,Mezura-Montes2011Constraint} are unable to directly process such a plethora of constraints, especially for a multi-objective context. 

One real-world application of such a problem is a robust landing trajectory design of supersonic transport (SST, i.e., civilian supersonic aircraft), where the uncertain vicissitude of wind or aerodynamic properties can severely influence its flight control. 

\subsection{Contribution}
In this paper, we propose an integrated method based on a constrained multi-objective evolutionary algorithm (MOEA) and non-intrusive polynomial chaos expansion (PCE) to solve a robust multi-objective constrained trajectory optimization problem. PCE is a widely used method for uncertainty quantification (UQ), which approximates an uncertainty by mapping known orthogonal basis polynomials.
By separating the constraints processed at the time of population generation, trajectory generation, and evaluation by MOEA, efficient constraint handling is achieved. 
Note that we only focus on the open-loop control in the present study. 

The contribution of this paper is threefold. 
First, we set up a robust multi-objective trajectory optimization problem with the generalized constraints, and rewrite this into a deterministic form with the PCE, keeping its mathematical rigorousness; the robust formulation developed in \cite{Li2014PCE_RobTrajOpt} is expanded to a multi-objective problem.
Secondly, we develop the constraint handling scheme through the unique integration of the formulated problem into the constrained MOEA, which the existing nonlinear programming (NLP)-based methods can no longer handle in the multi-objective case. This method can be applied to various multi-objective optimal control problems without any additional modifications.
Finally, we apply the method to the trajectory design in a preliminary vehicle design context, providing innovative insights into the robust landing control of the SST. 

\subsection{Related Work}

The robust optimization methods have been widely used \cite{Park2006RobustDesign, Du1999Robust}, as consideration of stochastic scenarios is vital for pragmatic engineering design. The complexity increases when changing the problem from a single-objective to a multi-objective case \cite{Deb2006RobustMO, ide2016robustness, gaspar2008robustness}. A dynamic optimization problem (DOP) considers time-variable parameters instead of the autonomous dynamical system (i.e., equations of motion). Even though this is generally categorized as a different problem from the trajectory optimization, several studies have addressed the robust multi-objective solution sets \cite{GUO2019MOO,Xin2010RobustOpt,Guo2018MOvehicle,jin2013framework}.

Some studies of (multi-objective) evolutionary computation have adopted the PCE to answer the uncertainty, with applications to an inverse problem \cite{Ho2012Inverse}, a layout of evacuation exits \cite{xie2018evacuation}, aerodynamic shape optimization \cite{liatsikouras2019aerodynamic, palar2015aerodyn, dodson2009aerodyn}, and life-cycle reliability design optimization \cite{ye2020pce}. However, these studies did not address the problems with an autonomous dynamical system, or the optimal control problem. 

In the field of trajectory optimization or path planning, numerous methods have been proposed to address uncertainty. The UQ \cite{Luo2017survey} has been adopted to transcribe a probabilistic expression into deterministic (thus tractable) forms, and the utilization of PCE \cite{Xiu2002PC, Eldred2008gPC} has been demonstrated to be a convincing method among other techniques, such as linear covariance analysis \cite{Saunders2012OptTraj, Tang2007rendevous}, Monte Carlo method \cite{Xu2014MC}, or ensemble prediction systems \cite{Gonzalez2018RobustAircraft}, due to its high expressiveness and mathematical rigorousness \cite{Li2014PCE_RobTrajOpt, Matsuno2015conflict, Fisher2011OptTrajGPC, Xiong2015PCE, Wang2019RobustConvex, Nakka2019Chanceconstr}. A limitation of these studies is their focus on single-objective problems. In particular, the integration of NLP-based optimization methods, which are the primary methods for single-objective optimal control problems, into a multi-objective problem is nontrivial.

In the aerospace field, the multi-objective trajectory optimizations are performed from the context of air traffic management \cite{Gardi2016ATM_MO}, vehicle design \cite{Fujikawa2015MDO} or interplanetary trajectory design \cite{englander2012mga}.
The direct predecessor of this study is \cite{Kanazaki2021SST}, which optimized the continuous descent operation (CDO) of the SST using the MOEA. The high-fidelity aerodynamic analysis was approximated by a surrogate model called Kriging model so that the efficient optimization was achieved. 

Some studies have solved multi-objective trajectory optimization problems robustly. 
Marto and Vasile proposed a method for the robust many-objective problem with epistemic uncertainty using surrogate modeling, where the objective was lower expectation \cite{Marto2021MOtraj}. However, if we are more interested in the probability distributions, the solutions will be too conservative. 
References \cite{Luo2007MOrendezvous, Luo2014robust, Yang2017robust, Gonzalez2019Aircraft} defined the problem-specific indices of robustness, which was regarded as one objective. However, such tailor-made ``robustness metrics" are difficult to generalize and apply directly to other problems. 
A multi-objective aircraft path planning problem with wind uncertainty was solved in \cite{Gonzalez2019Aircraft}, however the proposed algorithm did not include a general treatment of the constraints. In particular, path constraints were not considered in the paper. 

\subsection{Paper Organization}
The remainder of this paper is organized as follows. Section \ref{Methodology} defines the generalized robust multi-objective trajectory design problem and demonstrates the integrated optimization scheme. The SST landing trajectory design problem is introduced in Sec. \ref{SST_landing_formulation}. The results of the optimized trajectory are examined in Section \ref{results}, and Section \ref{Conclusion} concludes this paper. 

\section{Methodology}  \label{Methodology}

\subsection{Robust Multi-objective Trajectory Design Problem}  \label{ProbDef}

We start by setting up the robust multi-objective optimization problem, such as a trajectory optimization, as follows: 
\begin{subequations}
\begin{align}
\label{sto}
\min_{\mathbf{u}(t)} \quad & \{ \mathcal{J}_1, \ldots, \mathcal{J}_n \} \\ 
\label{sto_c1}
\text{subject to} \quad & \dot{\mathbf{x}}(t,\boldsymbol{\xi}) = \boldsymbol{F}(\mathbf{x}(t,\boldsymbol{\xi}), \mathbf{u}(t), t) \text{ a.s.}  \\
\label{sto_c2}
&\boldsymbol{g}(\mathbf{x}(t,\boldsymbol{\xi}), \mathbf{u}(t), t) \leq \boldsymbol{0} \text{ a.s.} \\
\label{sto_c3}
&\boldsymbol{h}(\mathbf{x}(t,\boldsymbol{\xi}), \mathbf{u}(t), t) = \boldsymbol{0} \text{ a.s.} 
\end{align}
\end{subequations}

\noindent where $\mathcal{J}_N, \forall N = 1, \dots n$ is an objective function; $\mathbf{x}(t)$ is a vector whose elements are state variables and $\mathbf{u}(t)$ is a vector whose elements are control variables to be optimized; and $\boldsymbol{\xi}$ is a probabilistic parameter, which brings uncertainty in the system. The objective of the optimal control problem is generalized as follows: 
\begin{equation}
    \mathcal{J}_N = \Phi (\mathbf{x}(0), \mathbf{u}(0), \mathbf{x}(t_f, \boldsymbol{\xi}), \mathbf{u}(t_f)) 
                    +
                    \int_{0}^{t_f} L (\mathbf{x}(t,\boldsymbol{\xi}), \mathbf{u}(t), t) \mathrm{d} t
\end{equation}

The constraint \eqref{sto_c1} represents the dynamics of the system, \eqref{sto_c2} is a set of inequality constraints (e.g., path constraints), and \eqref{sto_c3} is a set of equality constraints (e.g., initial and final constraints). 

Here, ``a.s." (``almost surely") enforces the robustness, thereby making the constraints stochastic. Note that $\mathbf{u}(t)$ does not include the uncertainty $\boldsymbol{\xi}$, as we consider the open-loop robust optimizer. 

\subsection{Reformulation to a Deterministic Problem}

In this subsection, we convert the robust dynamic optimization problem into an equivalent deterministic form, based on \cite{Li2014PCE_RobTrajOpt}.
First, we convert the stochastic static constraints into a set of equivalent deterministic constraints as follows: 
\begin{subequations}
\label{robust1}
\begin{align}
\label{robust1_obj}
\min_{\mathbf{u}(t)} \quad & \{ \mathcal{J}_1, \ldots, \mathcal{J}_n \} \\ 
\text{subject to} \quad & \label{robust1_c1}
\dot{\mathbf{x}}(t,\boldsymbol{\xi}) = \boldsymbol{F}(\mathbf{x}(t,\boldsymbol{\xi}), \mathbf{u}(t), t) \text{ a.s.}  \\
&\label{robust1_c2} 
\mu(\boldsymbol{g}(\mathbf{x}(t,\boldsymbol{\xi}), \mathbf{u}(t), t)) \leq \boldsymbol{0} \\
&\label{robust1_c3}
\sigma(\boldsymbol{g}(\mathbf{x}(t,\boldsymbol{\xi}), \mathbf{u}(t), t)) \leq \boldsymbol{\epsilon}_{g} \\
&\label{robust1_c4}
\mu( \boldsymbol{h}(\mathbf{x}(t,\boldsymbol{\xi}), \mathbf{u}(t), t)) = \boldsymbol{0}  \\
&\label{robust1_c5}
\sigma(\boldsymbol{h}(\mathbf{x}(t,\boldsymbol{\xi}), \mathbf{u}(t), t)) \leq \boldsymbol{\epsilon}_{h} 
\end{align}
\end{subequations}

\noindent This way, the probabilistic expression ``a.s." in \eqref{sto_c2} and \eqref{sto_c3} can be removed. This formulation is useful and applicable to various robust dynamic optimizations because the robustness of the terminal and path constraints is expressed in terms of the mean and standard deviation, which enables the designers to tune the problem parameters easily. 

Evidently, the stochastic description still remains in \eqref{robust1_c1}. To convert the robust dynamic constraint to a set of deterministic constraints, we introduce a non-intrusive PCE \cite{Eldred2008gPC}. 

When the dynamical system is nonlinear, an analytical solution of \eqref{robust1_c1} over a probabilistic distribution is rarely obtained. 
Instead, we consider an approximation for the state variable $\mathbf{x}(t,\boldsymbol{\xi})$ as a linear combination of orthogonal basis functions $\phi(\boldsymbol{\xi})$ with corresponding coefficients $\tilde{\boldsymbol{x}}(t)$. 
\begin{align}
    \mathbf{x}(t, \boldsymbol{\xi}) \simeq \sum_{i = 0}^{p} \tilde{\boldsymbol{x}}_{i}(t)\phi_i(\boldsymbol{\xi})
\end{align}

\noindent where $i$ is the order of approximation; in this equation, a $p$-th approximation is considered. The orthogonal basis functions have the following properties.
\begin{align}
\begin{split}
\label{orthogonality}
    <\phi_i(\boldsymbol{\xi}), \phi_j(\boldsymbol{\xi})> =  
    & \int \phi_i(\boldsymbol{\xi})\phi_j(\boldsymbol{\xi}) \rho(\boldsymbol{\xi})d \boldsymbol{\xi} \\ 
    = & <\phi_i^2>\delta_{ij}
\end{split}
\end{align}

\noindent where $\delta_{ij}$ is the Kronecker delta. Note that the combinations between the probability distribution of $\boldsymbol{\xi}$ and the pair of the orthogonal basis function $\phi(\boldsymbol{\xi})$ and corresponding weight (i.e., joint probability density) function $\rho(\boldsymbol{\xi})$ are predefined \cite{Xiu2002PC}. 

Based on the non-intrusive PCE, we have the following equation.
\begin{align} \label{non-intrusivePCE}
    \tilde{\boldsymbol{x}}_{i}(t) = \frac{1}{<\phi_i^2(\boldsymbol{\xi})>} \int \boldsymbol{x}(t, \boldsymbol{\xi}) \phi_i(\boldsymbol{\xi}) \rho(\boldsymbol{\xi}) d\boldsymbol{\xi}
\end{align}

\noindent Using the tensor-product quadrature rule, \eqref{non-intrusivePCE} can be approximated as follows:
\begin{align}
\label{quadrature}
    \tilde{\boldsymbol{x}}_{i}(t) = \sum_{s_1 = 1}^{l} \cdots \sum_{s_q = 1}^{l} \mathbf{x}(t,\xi_{s_1},\ldots,\xi_{s_q}) \frac{\phi_i(\xi_{s_1},\ldots,\xi_{s_q})}{<\phi_i^2>} \prod_{j = 1}^{q} w_j
\end{align}

\noindent where $\boldsymbol{\xi}=[\xi_{s_1},\ldots,\xi_{s_q}]$; $l$ is the number of integral points taken in the given range of the variables; $q$ is the dimensionality of vector $\boldsymbol{\xi}$ (i.e., number of uncertain factors considered); and $w_j$ indicates the weight of each integration point. 

The only unknown part in the right hand side of \eqref{quadrature} is $\mathbf{x}(t,\xi_{s_1},\ldots,\xi_{s_q})$. However, as all stochastic augments in this term are fixed, this can be solved as a deterministic ordinary differential equation, given an initial condition. 
\begin{align}
\begin{split}
\dot{\mathbf{x}}(t,\xi_{s_1},\ldots,\xi_{s_q}) &= \boldsymbol{F}(\mathbf{x}(t,\xi_{s_1},\ldots,\xi_{s_q}), \mathbf{u}(t), t)
\end{split}
\end{align}

\noindent Therefore, the constraint \eqref{robust1_c1} can be rewritten as follows: 
\begin{align}
\dot{\mathbf{x}}(t,\boldsymbol{\xi}_k) = \boldsymbol{F}(\mathbf{x}(t,\boldsymbol{\xi}_k), \mathbf{u}(t), t) \quad k = 1, \ldots, M
\end{align}

\noindent where $M$ is the number of the combinations of the vector $\boldsymbol{\xi}$, which is required by \eqref{quadrature} based on the PCE and quadrature rule: $M = l^q$. We thereby substitute one stochastic constraint \eqref{robust1_c1} with $M$ deterministic constraints, while maintaining the robustness. 

To summarize, the robust trajectory optimization problem \eqref{robust1} can be rewritten in a deterministic form.
\begin{subequations}
\label{robust2}
\begin{align}
\label{robust2_obj}
\min_{\mathbf{u}(t)} \quad & \{ \mathcal{J}_1, \ldots, \mathcal{J}_n \} \\
\text{subject to} \quad & \label{robust2_c1}
\dot{\mathbf{x}}(t,\boldsymbol{\xi}_k) = \boldsymbol{F}(\mathbf{x}(t,\boldsymbol{\xi}_k), \mathbf{u}(t), t) \quad k = 1, \ldots, M \\
&\label{robust2_c2} 
\mu(\boldsymbol{g}(\mathbf{x}(t,\boldsymbol{\xi}), \mathbf{u}(t), t)) \leq \boldsymbol{0} \\
&\label{robust2_c3}
\sigma(\boldsymbol{g}(\mathbf{x}(t,\boldsymbol{\xi}), \mathbf{u}(t), t)) \leq \boldsymbol{\epsilon}_{g} \\
&\label{robust2_c4}
\mu( \boldsymbol{h}(\mathbf{x}(t,\boldsymbol{\xi}), \mathbf{u}(t), t)) = \boldsymbol{0}  \\
&\label{robust2_c5}
\sigma(\boldsymbol{h}(\mathbf{x}(t,\boldsymbol{\xi}), \mathbf{u}(t), t)) \leq \boldsymbol{\epsilon}_{h} 
\end{align}
\end{subequations}

\noindent The reformulation of the stochastic dynamic constraint is generalized in \cite{Gonzalez2018RobustAircraft} as a ``trajectory ensemble," while that of the static constraints is different because they are shown as statistical properties here.

The mean and standard deviation of the state variables can be computed by the following formula, using  $\tilde{\boldsymbol{x}}_{i}(t)$.
\begin{align}
\label{mean_formula}
\mu (\mathbf{x}(t,\boldsymbol{\xi})) = & \tilde{\boldsymbol{x}}_{0}(t) \\
\label{std_formula}
\sigma (\mathbf{x}(t,\boldsymbol{\xi})) = &\sqrt {\sum_{i=1}^{p} <\phi_i^2> \tilde{\boldsymbol{x}}_{i}^2(t)}
\end{align}

\noindent These expressions can also be utilized to express the chance constraints via moment-based reformulation \cite{Muhlpfordt2019Chance_pce}.

\subsection{Optimization Method - Integration to MOEA}
\begin{figure}[t]
\centerline{\includegraphics[width=0.85\linewidth]{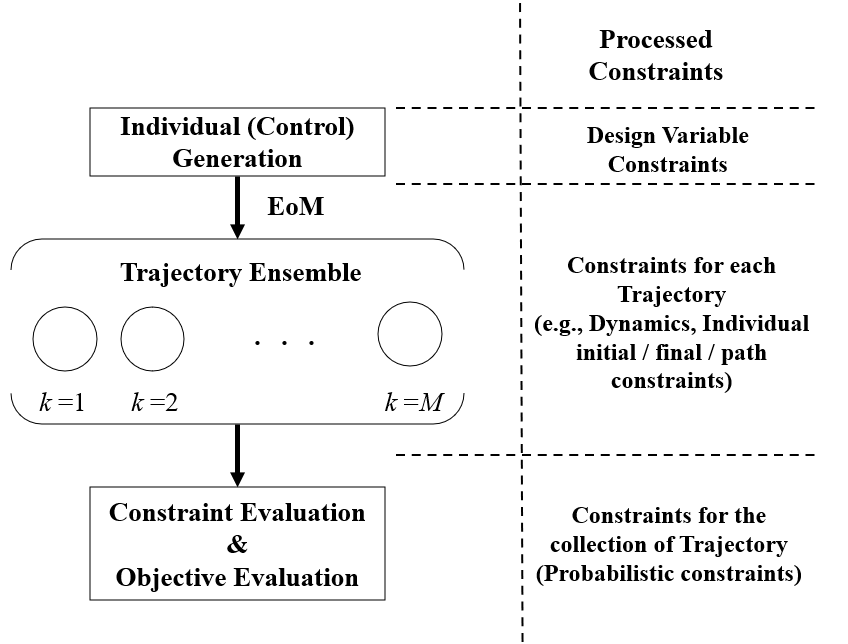}}
\caption{Constraint-handling scheme and integration of trajectory ensemble to the EA.}
\label{constraint_allocation}
\end{figure}

The robust multi-objective trajectory optimization problem \eqref{robust2} is solved using a constrained evolutionary algorithm (EA)\cite{coello2002theoretical, Deb2000CNSGA}. The constraint handling scheme and integration of the trajectory ensemble are shown in Fig. \ref{constraint_allocation}. The $M$ dynamic constraints shown in \eqref{robust2_c1} indicate different uncertain scenarios; given the same control input sequence (i.e., chromosome), $M$ distinct trajectories are generated. Then, one fitness function is returned by summing up the objectives of each trajectory.

In the proposed optimization method, the constraints are processed in three phases. First, when the individuals in a population are generated, the constraints can be enforced so that the design (control) variables are in a certain range. Next, the technique of trajectory ensemble guarantees that the individual constraints of each trajectory are automatically satisfied when the equations of motion (EoMs) are propagated; this corresponds to the constraints that are dealt with in the trajectory ensemble in \cite{Gonzalez2018RobustAircraft}. Finally, we can evaluate the statistical properties over the collection of trajectory ensembles by introducing the constrained MOEA; the information of the trajectories is sent to the EA, which handles the constraints in a statistical form while efficiently optimizing the multi-objective problem. In this way, we can optimize the same robust formulation of \cite{Li2014PCE_RobTrajOpt}, which processed all constraints at once by solving a large-scale NLP. 

\section{Application: SST Landing Trajectory Optimization}  \label{SST_landing_formulation}

As an example application of the developed method to real-world problems, in this section we optimize a robust landing trajectory of an SST. 

During the preliminary design of aerospace vehicles, their trajectories are oftentimes coupled with the objective functions to be optimized. Hence, the necessity for an efficient optimization method for the flight control system arises. Moreover, the aerodynamic properties of supersonic vehicles has a large impact on its optimal trajectories, thereby requiring a coupled analysis of these two factors. \cite{Kanazaki2021SST, Fujikawa2015MDO}.

However, delta wings with a low aspect ratio (as shown in Fig. \ref{SST}) are usually selected for the wing shape of SSTs to maximize their performance in a supersonic domain, which results in limited lift being available in a subsonic domain \cite{Razak1966}. Thus, the uncertain factors, such as the vicissitude of the aerodynamics or the meteorological conditions, impose non-negligible variation when performing a low-speed trajectory optimization. This motivates robustness in the flight control during landing and takeoff, even for the open-loop control. 

\subsection{Deterministic Problem Formulation}
\begin{figure}[tb]
\centerline{\includegraphics[scale=0.4]{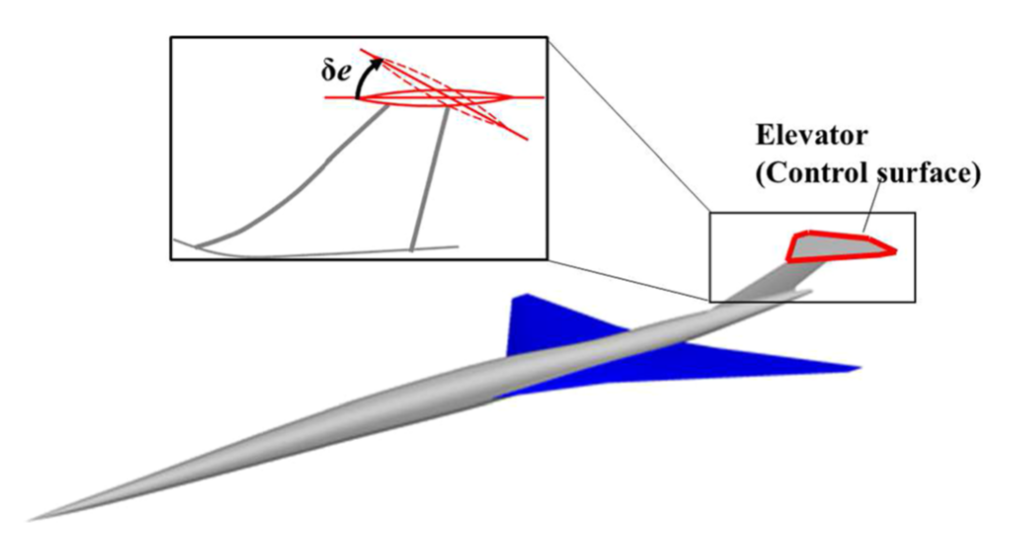}}
\caption{Simplified SST model used in this study \cite{Kanazaki2021SST}.}
\label{SST}
\end{figure}

The baseline deterministic formulation of the landing trajectory optimization problem based on \cite{Kanazaki2021SST} is shown in \eqref{det_SST}.
The control input of the elevator $\delta e$ is optimized during the CDO.
The SST model used in this problem is shown in Fig. \ref{SST}. For simplicity, the effect of engine nacelles installations is not considered; the thrust is also ignored so that there is only one control variable, making the comparison between the robust control and the nominal control clearer. However, the inclusion of these aspects in the proposed formulation is trivial.  
\begin{subequations}
\label{det_SST}
\begin{align}
\max_{\delta e(t)} \quad & \{t_f, x(t_f)\} \\
\text{subject to} \quad \nonumber\\
\label{det_SST_c1}
\dot{\mathbf{x}}(t) = &
\left[\begin{array}{c}
\dot{x} \\
\dot{z} \\
\dot{u} \\
\dot{w} \\
\dot{\theta} \\
\dot{q} \\
\end{array}\right] =
\left[\begin{array}{c}
u \\
w \\
\frac{X}{m} - g \sin{\theta} - \frac{d\theta}{dt}\frac{dz}{dt}  \\
\frac{Z}{m} - g \cos{\theta} -  \frac{d\theta}{dt}\frac{dx}{dt} \\
q \\
\frac{M_\theta}{I_{yy}} - \frac{d\theta}{dt}  \\
\end{array}\right] \\
\label{det_SST_c2}
\SI{-50.0}{\degree} &\leq \delta e(t) \leq \SI{10.0}{\degree}, \ \SI{-35.9}{\degree} \leq \delta e(0) \leq \SI{-15.9}{\degree}\\
\label{det_SST_c2_1}
|\delta e(t+&1) - \delta e(t)| \leq 2.0 ^{\circ}\\
\label{det_SST_c3}
0.1 \leq & Ma(t) \leq 0.5 \\
\label{det_SST_c4}
-5.0 \leq & \alpha(t) \leq 21.0 ^{\circ} \\
\label{det_SST_c5} 
\mathbf{x}(0)& = \mathbf{x}_0 =
\left[\begin{array}{c}
x_0 \\
z_0 \\
u_0 \\
w_0 \\
\theta_0 \\
q_0 \\
\end{array}\right] =
\left[\begin{array}{c}
\SI{0.0}{\meter}\\
\SI{1000.0}{\meter} \\
\SI{120.0}{\meter/\second} \\
\SI{0.0}{\meter/\second} \\
\SI{0.0}{\degree} \\
\SI{0.0}{\degree/\second} \\
\end{array}\right] \\
\label{det_SST_c6} 
z(t_f) &= \SI{0.0}{\meter} 
\end{align}
\end{subequations}

\begin{figure}[tb]
\centerline{\includegraphics[scale=0.45]{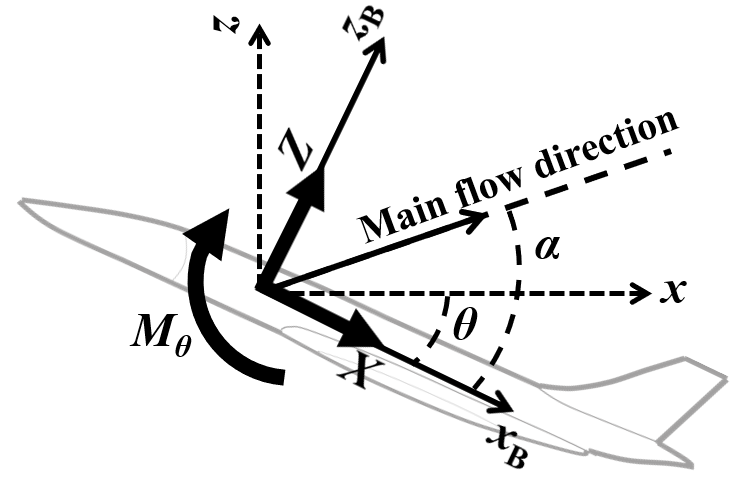}}
\caption{Coordinate frame of the SST \cite{Kanazaki2021SST}.}
\label{Coordinate}
\end{figure}

\noindent The coordinate system is presented in Fig. \ref{Coordinate}. In this problem, a 3 degrees-of-freedom (DoF) trajectory in the $x-z$ plane (inertial frame) is considered. $I_{yy}$ is the moment of inertia about $y$-axis; $m$ is the mass of the aircraft; $X$ and $Z$ in \eqref{det_SST_c1} are the aerodynamic forces along the body axes $x_B$ and $z_B$, respectively; and $M_\theta$ is the pitching moment based on the pitch angle $\theta$. $X,Z,$ and $M_\theta$ are estimated using the Kriging method, where the details are explained in Sec. \ref{kriging_}.

The upper and lower bounds, as well as the initial condition of the control input, are determined in \eqref{det_SST_c2}. Additionally, \eqref{det_SST_c2_1} ensures that the elevator angle does not move more than $2^{\circ}$ in one second. Additionally, this case study assumes that the elevator control is performed twice in each second. The constraints \eqref{det_SST_c3} and \eqref{det_SST_c4} are the path constraints, and the feasible velocity (in Mach number $Ma$) and angle of attack $\alpha$ bounded. The constraint \eqref{det_SST_c5} is the initial state of the vehicle, and \eqref{det_SST_c6} is the terminal condition, which is the completion of landing. 

\subsection{Robust Problem Formulation} \label{SST_robust_form}

Integrating the uncertainty $\boldsymbol{\xi}$ into \eqref{det_SST}, the problem can be rewritten in the form of \eqref{robust2}. 
\begin{subequations}
\label{robust_SST}
\begin{align}
\max_{\delta e(t)} \quad & \{\mu(t_f(\boldsymbol{\xi})), \mu(x(t_f,\boldsymbol{\xi}))\} \\
\text{subject to} \quad \nonumber\\
\label{robust_SST_c1}
\dot{\mathbf{x}}(t,\boldsymbol{\xi}_k)& = \boldsymbol{F}(\mathbf{x}(t,\boldsymbol{\xi}_k), \delta e(t), t) \quad k = 1, \ldots, M \\
\label{robust_SST_c2}
\SI{-50.0}{\degree} &\leq \delta e(t) \leq \SI{10.0}{\degree}, \ \SI{-35.9}{\degree} \leq \delta e(0) \leq \SI{-15.9}{\degree}\\
\label{robust_SST_c2_1}
|\delta e(t+&1) - \delta e(t)| \leq \SI{2.0}{\degree}\\
\label{robust_SST_c3}
0.1 \leq & \mu(Ma(t,\boldsymbol{\xi})) \leq 0.5 \\
\label{robust_SST_c4}
& \sigma(Ma(t,\boldsymbol{\xi}))\leq 0.2 \\
\label{robust_SST_c5}
\SI{-5.0}{\degree} \leq & \mu(\alpha(t,\boldsymbol{\xi})) \leq \SI{21.0}{\degree} \\
\label{robust_SST_c6}
& \sigma(\alpha(t,\boldsymbol{\xi}))\leq \SI{4.0}{\degree} \\
\label{robust_SST_c7}
& \mathbf{x}(0, \boldsymbol{\xi}) = \mathbf{x}_0 \\
\label{robust_SST_c8}
& z(t_f,\boldsymbol{\xi}) = \SI{0.0}{\meter} \\
\label{robust_SST_c9}
& \sigma(t_f,\boldsymbol{\xi}) \leq \sigma_1 \\
\label{robust_SST_c10}
& \sigma(x(t_f,\boldsymbol{\xi})) \leq \sigma_2
\end{align}
\end{subequations}

\noindent The averages of each objective over the uncertainty are chosen as the new objectives of \eqref{robust_SST}. To guarantee the robustness of the objectives, two additional constraints, \eqref{robust_SST_c9} and \eqref{robust_SST_c10}, impose the terminal standard deviation so that the distributions of the objectives are guaranteed to be in a certain (predefined) range.
The path constraints \eqref{det_SST_c3} and \eqref{det_SST_c4} are now split into inequalities of the mean and standard deviation, \eqref{robust_SST_c3} - \eqref{robust_SST_c6}. 

Because the constraints \eqref{robust_SST_c2} and \eqref{robust_SST_c2_1} are the upper and lower bounds of the elevator control angle, which is physically impermissible to infringe, they are kept in their deterministic forms. Note that a robust control input is invariant to the uncertainty, and thus its standard deviation over the uncertainty is always zero. Furthermore, the terminal constraints \eqref{robust_SST_c8} and \eqref{robust_SST_c7} are kept deterministic by the definition of the problem.

Furthermore, when the constrained MOEA processes the constraints shown in \eqref{robust_SST_c4}, \eqref{robust_SST_c6}, \eqref{robust_SST_c9} and \eqref{robust_SST_c10}, the values of the standard deviation are approximated by \eqref{std_formula}. Even though some variables are not the elements of $\mathbf{x}(t)$, they can be derived from the state variables easily.

In this case study, an uncertain wind velocity in the horizontal  ($x$-) direction is modeled as an uniform distribution in the range $\xi \in [-5,5] \SI{}{\meter/\second}$ (i.e., $q=1$); the corresponding orthogonal basis function for this probability distribution is a Legendre polynomial \cite{Xiu2002PC}. Here, we assume that the wind velocity stays constant during the flight.

\subsection{Efficient Aerodynamic Estimation with the Kriging Method}
\label{kriging_}
The aerodynamic forces and moment in \eqref{det_SST_c1} and \eqref{robust_SST_c1} are expressed as follows:
\begin{align}
\label{aero_forces}
& X = X_0 + Q \cdot \delta e \cdot  C_{X\delta e} (Ma, \alpha, \delta e) \\
& Z = Z_0 + Q \cdot \delta e  \cdot  C_{Z\delta e} (Ma, \alpha, \delta e) \\
& M_\theta = M_{\theta0} + Q  \cdot \delta e  \cdot C_{M_\theta \delta e} (Ma, \alpha, \delta e) 
\end{align}

\noindent where $X_0$, $Z_0$ and $M_{\theta0}$ are the aerodynamic forces and moments when the elevator is not controlled; $Q$ is the dynamic pressure; and $C_{X\delta e}$, $C_{Z\delta e}$, and $C_{M_\theta\delta e}$ are sensibility coefficients of the elevator control, which are the functions of Mach number, angle of attack, and elevator control. The values of these coefficients are estimated via semi-empirical aerodynamic prediction based on the Kriging model. 

The Kriging method is a spatial domain interpolation technique that estimates the value of the unsampled points with an assumption of a Gaussian process \cite{Jones1998Kriging}. The value of any data point $f(\boldsymbol{\chi})$ is approximated as follows:
\begin{equation}
    f(\boldsymbol{\chi}) = \mu + \epsilon(\boldsymbol{\chi})
\end{equation}

\noindent where $\mu$ is the average of all sampled points (and therefore constant), and $\epsilon(\boldsymbol{\chi})$ is a local deviation function from the average. In the Kriging model, this local deviation function learns the offset of $f(\boldsymbol{\chi})$ from the average and the credibility of that offset as a form of covariance, assuming a Gaussian distribution. This stochasticity contributes to the accuracy of the model, compared to learning only the mean offset. 

The aerodynamic coefficients of the sample points are obtained by computational fluid dynamics (CFD) analysis, which numerically solves the Reynolds-averaged Naiver-Stokes (RANS) equation \cite{Kanazaki2021SST}. The precomputation of this the data set enables us to avoid running the high-fidelity but time-consuming CFD simulation for each time step, while still obtaining the aerodynamic properties efficiently. The sample points collected to build the model are listed in Table \ref{kriging}. 

\begin{table}[htbp]
\caption{Sample points of Kriging model for the aerodynamic database}
\label{algo_para}
\begin{center}
\begin{tabular}{|c|c|}
\hline
\textbf{Parameter}&\textbf{Sample Points} \\
\hline
$Ma$ & $0.1, 0.3, 0.5$  \\
$\alpha$ & $-5^{\circ} \leq \alpha \leq 21^{\circ}$ (every $2^{\circ}$)\\
$\delta e$ &  $-50^{\circ} \leq \delta e \leq 10^{\circ}$ (every $5^{\circ}$)\\
\hline
\end{tabular}
\label{kriging}
\end{center}
\end{table}

\subsection{Integration with the Proposed Optimization Methods} \label{integration}

\begin{figure}[tb]
\centerline{\includegraphics[scale=0.5]{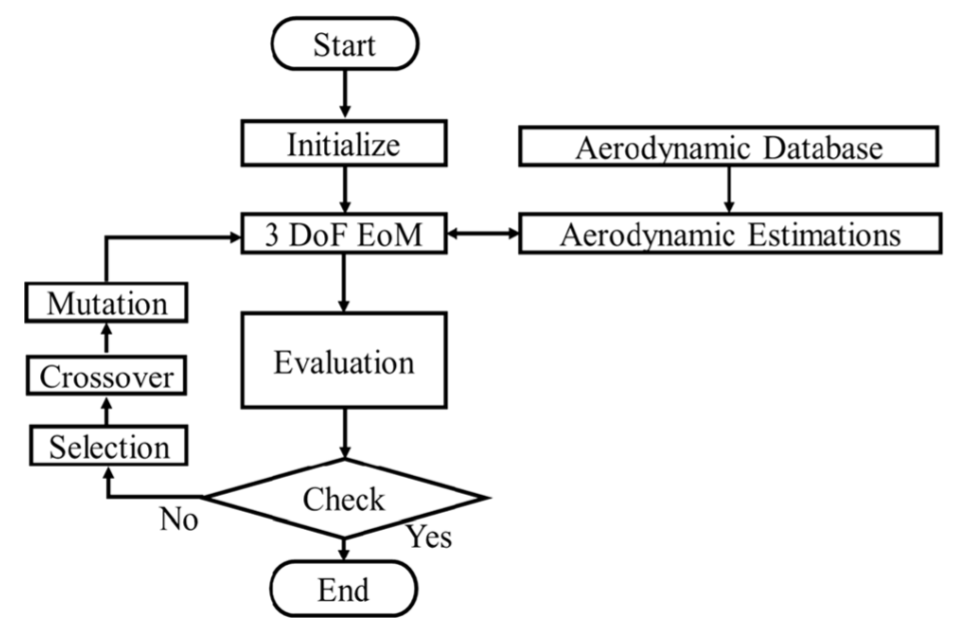}}
\caption{Optimization architecture of trajectory design with high-fidelity aerodynamic estimation \cite{Kanazaki2021SST}.}
\label{Procedure}
\end{figure}

The integrated MOEA architecture for this problem is presented in Fig. \ref{Procedure}. The chromosome for each individual represents a control history. At each time step of this propagation, the parameters (i.e., $Ma, \alpha$, and $\delta e$) are passed to the Kriging model, which approximates the sensitivities of the elevator control. These values are returned to the EoMs with the control input, and the states are propagated by one time step. By repeating this process, we can obtain the flight trajectories, which are then evaluated by the MOEA.

An issue specific to this case study is that, when solving the robust optimization problem, \eqref{robust_SST_c3} and \eqref{robust_SST_c5} allow each state variable to exceed the range of the Kriging model for aerodynamic estimations. To avoid infeasibility, we tighten these statistical constraints to individual ones so that each trajectory satisfies the original path constraints \eqref{det_SST_c3} and  \eqref{det_SST_c4}. This reformulation enables us to process these constraints during trajectory generation, which reduces the number of constraints dealt with during constraint evaluation by the constrained EA. 

\begin{figure}[tb]
\centerline{\includegraphics[width=0.85\linewidth]{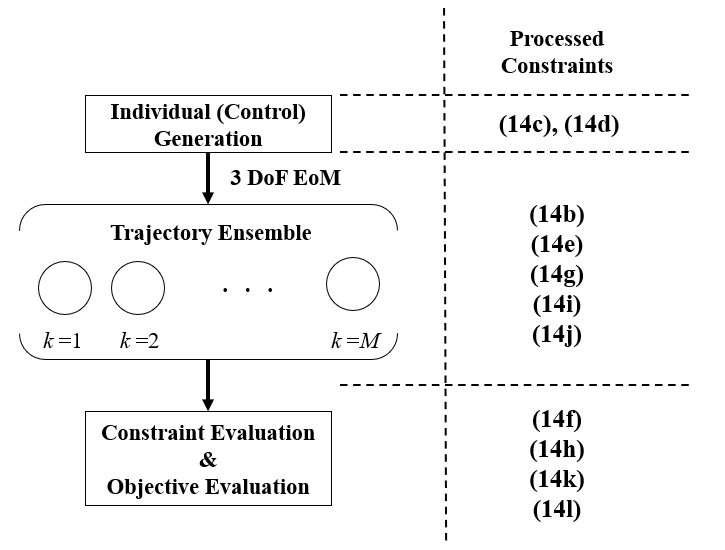}}
\caption{Constraint-handling of the robust formulation \eqref{robust_SST}.}
\label{constraints_process}
\end{figure}

The treatment of the constraints in the proposed algorithm is shown in Fig. \ref{constraints_process}. The control constraints \eqref{robust_SST_c2} and \eqref{robust_SST_c3} are considered when the population is generated, and the design variables are clipped to the upper and lower bounds. As the 3 DoF EoMs are solved numerically, each trajectory in the ensemble automatically satisfies \eqref{robust_SST_c1}, \eqref{robust_SST_c7}, and \eqref{robust_SST_c8}. As mentioned above, the path constraints \eqref{robust_SST_c3} and \eqref{robust_SST_c5} are individually checked at each time step in this case study; if these constraints are violated, the numerical calculation of the EoMs is terminated, and information regarding the unsatisfaction of the constraint(s) is sent to the constrained EA. Finally, the standard deviations \eqref{robust_SST_c4}, \eqref{robust_SST_c6}, \eqref{robust_SST_c9}, and \eqref{robust_SST_c10} are evaluated by the constrained MOEA after the trajectory is completely generated. Although not adopted in this case study, the mean path constraints that have a form of \eqref{mean_formula} may be evaluated with the standard deviation path constraints in the same way.  

\subsection{Implementation}
For the constrained MOEA, the constrained Non-dominated Sorting Genetic Algorithm II (CNSGA-II) \cite{Deb2000CNSGA} with polynomial mutation and simulated binary crossover was adopted. The EoMs were numerically solved using the fourth-order Runge-Kutta method with a time step of $\Delta t = \SI{0.1}{\second}$. 

The number of integral points is set to $l=6$, and the order of the PCE is $p=4$, both of which are the same to \cite{Li2014PCE_RobTrajOpt}.

\section{Results and Analysis} \label{results} 

Both the deterministic problem \eqref{det_SST} and the robust problem \eqref{robust_SST} were solved, and the optimized solution sets are compared in this section. The number of individuals in each generation was set to 30, and 600 generations of evolution were performed.

Additionally, to see the solution response to different values of the robustness constraints, we set three constraint values for \eqref{robust_SST_c9} and \eqref{robust_SST_c10}, which are shown in Table \ref{Sigmas}.

\begin{table}[tb]
\caption{Constraints values of $\sigma_1$ and $\sigma_2$}
\label{Sigmas}
\begin{center}
\begin{tabular}{|c|c|c|}
\hline
\textbf{Values} & \boldmath{$\sigma_1$}, s & \boldmath{$\sigma_2$}, m  \\
\hline
Low standard deviation & 1.0 & 100.0 \\
Baseline & 3.0 & 150.0 \\
High standard deviation & 5.0 & 200.0 \\

\hline
\end{tabular}
\end{center}
\end{table}

\subsection{Noninferior Solutions Sets}

\begin{figure}[tb]
    \centering
  \subfloat[$\sigma_1$-dependency of the optimized solution sets \label{1a}]{%
       \includegraphics[width=0.85\linewidth]{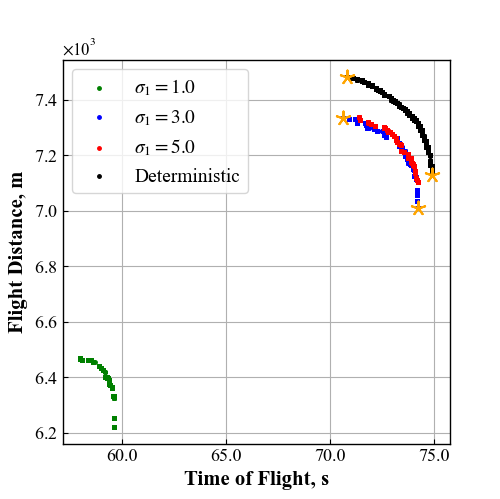}}
    \hfill
  \\
  \subfloat[$\sigma_2$-dependency of the optimized solution sets \label{1b}]{%
        \includegraphics[width=0.85\linewidth]{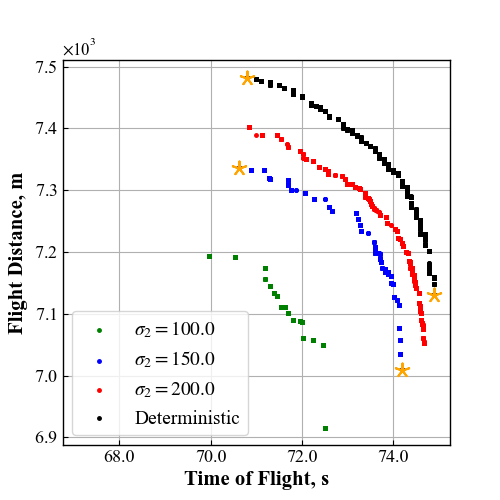}}
  \caption{Noninferiror solution sets for different stochastic scenarios. Orange stars indicate the representative solution points used in the trajectory analysis.}
  \label{noninf_plot} 
\end{figure}

The sets of noninferior solutions for each trial (i.e., three different sensitivity analysis cases and one deterministic case) are shown in Fig. \ref{noninf_plot}. For the robust solution points, the average of the objectives calculated using \eqref{mean_formula} are plotted. When varying $\sigma_1$ or $\sigma_2$, the other value was set to its baseline value. 

All noninferior solutions sets of the robust solutions were observed to be more conservative than those of the deterministic nominal solutions. A significant discrepancy was observed between $\sigma_1 = 1.0$ and $\sigma_1 = 3.0$, as shown in Fig. \ref{1a}. On the other hand, the solution plots and hypervolumes of $\sigma_1 = 3.0$ and $\sigma_1 = 5.0$ were almost identical, which indicates that the deviation of $\sigma_1$ was irrelevant to the noninferior solutions set in this context. However, the general trend was confirmed to be that the tightened standard deviation constraints (i.e., smaller values of $\sigma_1$ and $\sigma_2$) make the optimizer more conservative. Note that the conversion of \eqref{robust_SST_c3} and \eqref{robust_SST_c5} into an individual constraint, as discussed in Sec. \ref{integration}, is considered to have slightly contributed to making the solution sets conservative, although the sensitivity analysis shows that the original problem formulation itself makes the solution sets robust.



\subsection{Representative Trajectories}

\begin{figure}[tb]
    \centering
  \subfloat[$\sigma_1$-dependency, max. $t_f$\label{3a}]{%
       \includegraphics[width=0.95\linewidth]{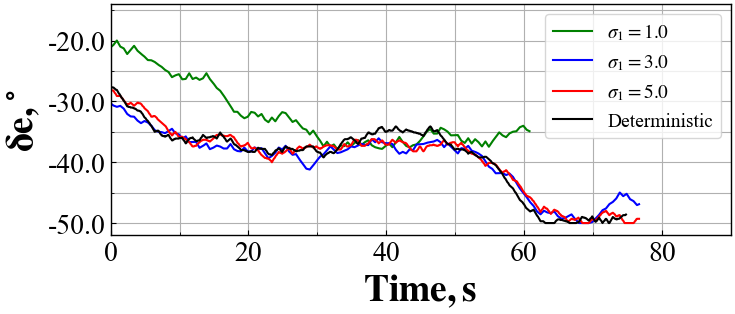}}
    \hfill
  \\
  \subfloat[$\sigma_1$-dependency, max. $x(t_f)$\label{3b}]{%
        \includegraphics[width=0.95\linewidth]{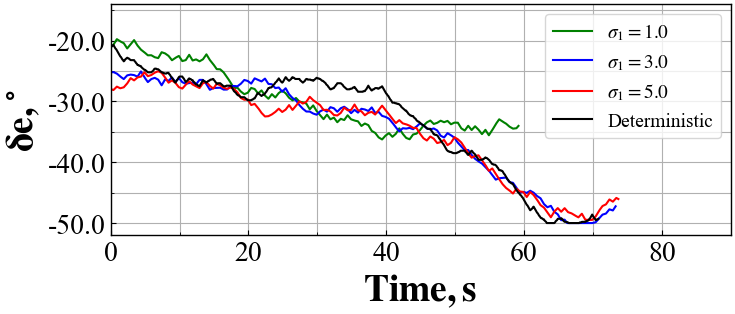}}
    \hfill
  \\
  \subfloat[$\sigma_2$-dependency, max. $t_f$\label{3c}]{%
       \includegraphics[width=0.95\linewidth]{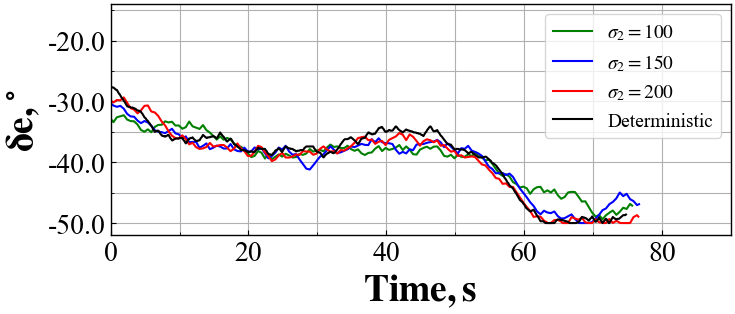}}
    \hfill
  \\
   \subfloat[$\sigma_2$-dependency, max. $x(t_f)$\label{3d}]{%
       \includegraphics[width=0.95\linewidth]{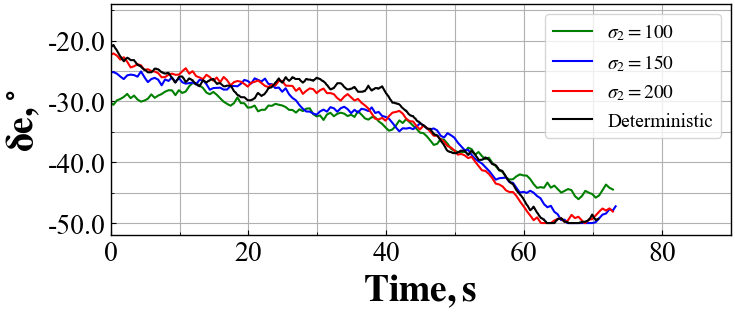}}
  \caption{Optimal control histories of extreme Pareto solutions.}
  \label{control_hist} 
\end{figure}

The representative control histories from the noninferior solution sets (extreme Pareto solutions) are shown in Fig. \ref{control_hist}. This figure shows the controls of the solution points that maximize $t_f$ and $x(t_f)$ among the noninferior solutions for each sensitivity analysis of $\sigma_1$ and $\sigma_2$. The plotted robust controls are terminated at the longest time of flight over the uncertainties. Except for the case of $\sigma_1=1.0$, most optimal control histories were identical regardless of the values of $\sigma$. However, different control strategies were observed for the maximization of each objective; Fig. \ref{3a} and \ref{3c} show that $\delta e$ decreased and then increased again before a sudden drop near the landing to maximize $t_f$, while a monotonic decrease of the elevator angle was preferred to maximize $x(t_f)$, as presented in Fig. \ref{3b} and \ref{3d}. 


\begin{figure} [tb]
    \centering
  \subfloat[max. $t_f$ solution \label{4a}]{%
       \includegraphics[width=1.0\linewidth]{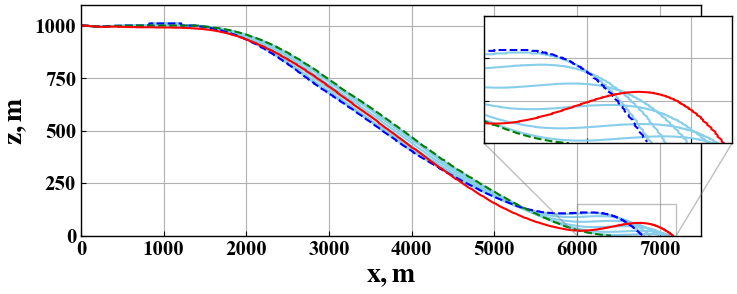}}
    \hfill
  \\
  \subfloat[max. $x(t_f)$ solution \label{4b}]{%
        \includegraphics[width=1.0\linewidth]{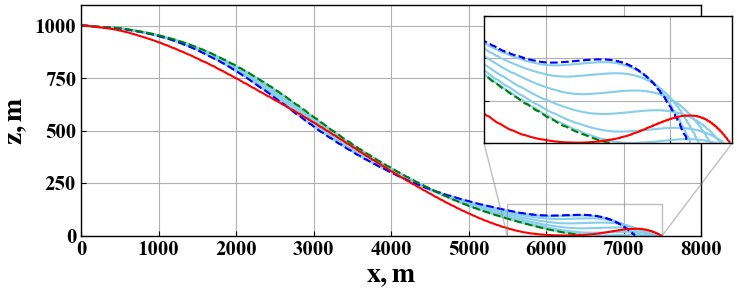}}
    \hfill
  \caption{Comparison of the nominal trajectory and baseline robust trajectory. Red lines indicate the nominal trajectories. Each sky blue trajectory represents a generated trajectory that corresponds to the integral points (trajectory ensemble), and dotted trajectories correspond to the maximum tailwind (green) and headwind (blue) scenarios.}
  \label{traj_hist} 
\end{figure}

The generated trajectories of the representative solutions are presented in Fig. \ref{traj_hist}. The deterministic solutions and the baseline robust solutions are compared; the maximum of both $t_f$ and $x(t_f)$ trajectories are presented, which are identified by the orange asterisks in Fig. \ref{noninf_plot}. In each subfigure, one trajectory from the deterministic solution point and eight trajectories from the robust solution point (six trajectories corresponding to each integration point, the worst scenario (i.e., maximum headwind), and the best scenario (i.e., maximum tailwind)), are plotted. 

Figs. \ref{4a} and \ref{4b} show that the nominal and robust trajectories started to differentiate below $z=\SI{250}{m}$, where the uncertainty had a notable influence on them. This could be because the velocity of the aircraft decreases as it approaches the ground, which provides a smaller lift, resulting in the leverage of the uncertainty increasing. 

\begin{figure} [tb]
    \centering
  \subfloat[max. $t_f$ solution \label{5a}]{%
       \includegraphics[width=1.0\linewidth]{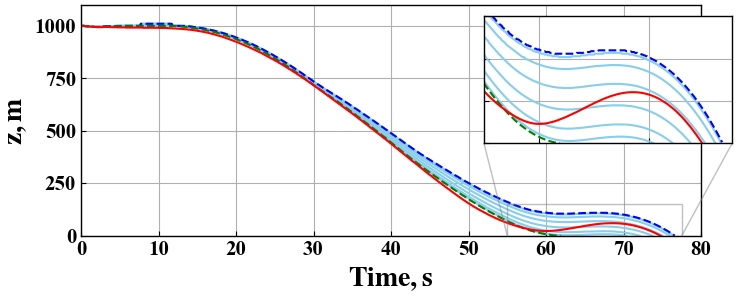}}
    \hfill
  \\
  \subfloat[max. $x(t_f)$ solution \label{5b}]{%
        \includegraphics[width=1.0\linewidth]{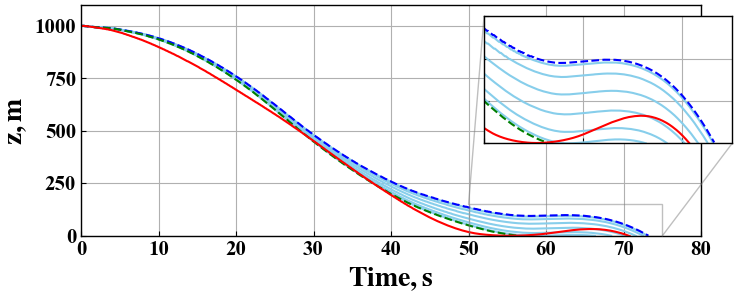}}
    \hfill
  \caption{Comparison of altitude history. Labeling is the same as in Fig. \ref{traj_hist}.}
  \label{tz_hist} 
\end{figure}

While both trajectories included a phase where the altitude was regained before the landing, these ascents were steeper in the nominal trajectories. For example, the nominal trajectory in Fig. \ref{4b} almost touched the ground before ascending around $x=\SI{6000}{\meter}$; this control would likely cause the early landing of the aircraft if it were to experience a strong headwind. On the other hand, the ascent in each sampled robust trajectory is much less aggressive, taking into account the multiple EoMs and avoiding the early landing in the case of the worst wind uncertainty. Such an observation validates the importance of the newly provided robust control. This can also be confirmed in the history of the altitude, presented in Fig. \ref{tz_hist}, as the robust trajectories almost always flew higher than the nominal trajectory to make the flight plan more robust. 

The optimized solutions were validated by a Monte Carlo (MC) test, using the two robust solution points: maximizers of $t_f$ and $x(t_f)$ with the baseline uncertainty. 1000 MC simulations were conducted, and the standard deviations of the two objectives were shown in Table \ref{MC}. The calculated $\sigma_1$ and $\sigma_2$ clearly satisfied the constraints \eqref{robust_SST_c9} and \eqref{robust_SST_c10}.

\begin{table}[t]
\caption{Terminal standard deviations of objectives based on Monte Carlo simulations}
\label{MC}
\begin{center}
\begin{tabular}{|l|c|c|}
\hline
\textbf{Solution points} & \boldmath{$\sigma_1$}, s & \boldmath{$\sigma_2$}, m  \\
\hline
Baseline, max. $t_f$ solution    & 2.02  & 117.38  \\
Baseline, max. $x(t_f)$ solution & 2.70  & 145.96  \\
\hline
\end{tabular}
\label{tab1}
\end{center}
\end{table}

An important observation is that the aircraft ``crashed" into the ground in the worst cases and failed to reraise its altitude, however the trajectories that correspond to the minimum value of the integral points succeeded in reraising, which extended the time of flight and flight distance. Even though the terminal state distributions of the representative solution points satisfied the statistical constraints, as shown in Table \ref{MC}, the worst scenario may still cause a significantly different outcome. This limitation is due to the nature of the robust formulation \eqref{robust1}, and further investigation into preventing the destructive worst scenario is left to future studies.


\section{Conclusion} \label{Conclusion}

Many real-world trajectory optimization problems are constrained, multi-objective, and uncertain. However, investigations into such problems are yet to be thoroughly conducted. The method proposed in this paper is a generalized formulation of robust multi-objective trajectory optimization based on the non-intrusive PCE, and the optimization method based on the constrained MOEA with the sequential constraint-handling scheme. These are applicable to any open-loop optimal control problem as long as the constraints of the state and control variables can be decoupled. 

This optimization architecture is applied to the landing trajectory design of the SST. This problem has an uncertainty in the wind velocity, and the robust optimal controls of the CDO are obtained by sensitivity analysis. The robust and conservative objective surfaces and trajectories are observed by limiting the standard deviation of the objectives over the uncertainty.

For future work, means to further accelerate the computational cost of the trajectory ensemble of the PCE should be considered, following research such as GPU parallelization \cite{Gonzalez2019Aircraft}, efficient sparse grid quadrature method \cite{Eldred2008gPC}, or the Kriging method \cite{Matsuno2016conflict}; the introduction of multi-fidelity methods is critical when increasing the dimension of the uncertainty. 
Furthermore, we expect that the proposed method will be applied not only to the other aerospace vehicles such as spacecraft rendezvous or Mars aircraft bust also to various engineering problems such as robotics or traffic flows, leveraging the generality of the problem formulation.  

\section*{Acknowledgment}

Yuji Takubo thanks Yusuke Yamada for many pieces of advice regarding the implementation of algorithms, and Thomas Van Buren for proofreading. 

\vspace{12pt}

\bibliographystyle{IEEEtran}
\bibliography{reference}

\end{document}